\documentclass[sigconf]{acmart}

\usepackage{amssymb}
\usepackage{booktabs}
\usepackage{float}
\usepackage{multirow}
\usepackage{stfloats}
\usepackage{colortbl}
\usepackage{tabularx}
\usepackage{pifont} 
\usepackage{bbding} 
\usepackage{fontawesome} 
\usepackage{ulem}

\newcommand{\cmark}{\ding{51}}
\newcommand{\xmark}{\ding{55}}

\AtBeginDocument{%
  }

\copyrightyear{2025}
\acmYear{2025}
\setcopyright{acmlicensed}
\acmConference[MM '25] {Proceedings of the 33rd ACM International Conference on Multimedia}{October 27--31, 2025}{Dublin, Ireland.}
\acmBooktitle{Proceedings of the 33rd ACM International Conference on Multimedia (MM '25), October 27--31, 2025, Dublin, Ireland}
\acmISBN{979-8-4007-2035-2/2025/10}
\acmDOI{10.1145/3746027.3758232}
\begin{document}

\title{MultiEgo: A Multi-View Egocentric Video Dataset for 4D Scene Reconstruction}


\settopmatter{authorsperrow=4}

\author{Bate Li}
\affiliation{%
  \institution{Shanghai Jiao Tong University}
  \state{Shanghai}
  \country{China}
}
\email{woxelikeloud@sjtu.edu.cn}

\author{Houqiang Zhong}
\affiliation{%
  \institution{Shanghai Jiao Tong University}
  \state{Shanghai}
  \country{China}
}
\email{zhonghouqiang@sjtu.edu.cn}

\author{Zhengxue Cheng}
\authornotemark[1]
\affiliation{%
  \institution{Shanghai Jiao Tong University}
  \state{Shanghai}
  \country{China}
}
\email{zxcheng@sjtu.edu.cn}

\author{Qiang Hu}
\affiliation{%
  \institution{Shanghai Jiao Tong University}
  \state{Shanghai}
  \country{China}
}
\email{qiang.hu@sjtu.edu.cn}

\author{Qiang Wang}
\affiliation{%
  \institution{VisionStar Information Technology (Shanghai) Co., Ltd.}
  \state{Shanghai}
  \country{China}
}
\email{wq@sightp.com}

\author{Li Song}
\authornotemark[1]
\affiliation{%
  \institution{Shanghai Jiao Tong University}
  \state{Shanghai}
  \country{China}
}
\email{song_li@sjtu.edu.cn}

\author{Wenjun Zhang}
\authornote{Corresponding authors.}
\affiliation{%
  \institution{Shanghai Jiao Tong University}
  \state{Shanghai}
  \country{China}
}
\email{zhangwenjun@sjtu.edu.cn}

\renewcommand{\shortauthors}{Bate Li et al.}

\begin{abstract}
  Multi-view egocentric dynamic scene reconstruction holds significant research value for applications in holographic documentation of social interactions. 
  However, existing reconstruction datasets focus on static multi-view or single-egocentric view setups, lacking multi-view egocentric datasets for dynamic scene reconstruction. 
  Therefore, we present MultiEgo, the \textbf{first} multi-view egocentric dataset for 4D dynamic scene reconstruction. The dataset comprises five canonical social interaction scenes: meetings, performances, and a presentation. Each scene provides five authentic egocentric videos captured by participants wearing AR glasses. We design a hardware-based data acquisition system and processing pipeline, achieving sub-millisecond temporal synchronization across views, coupled with accurate pose annotations. Experiment validation demonstrates the practical utility and effectiveness of our dataset for free-viewpoint video (FVV) applications, establishing MultiEgo as a foundational resource for advancing multi-view egocentric dynamic scene reconstruction research.
  Our project page and dataset are available at \url{https://woxelikeloud.github.io/multiego/}.
\end{abstract}

\ccsdesc[500]{Computing methodologies~Computer graphics}
\ccsdesc[500]{Computing methodologies~Computer vision}


\keywords{Egocentric Video; Free-viewpoint Video; Dynamic Scene }
\begin{teaserfigure}
  \includegraphics[width=\textwidth]{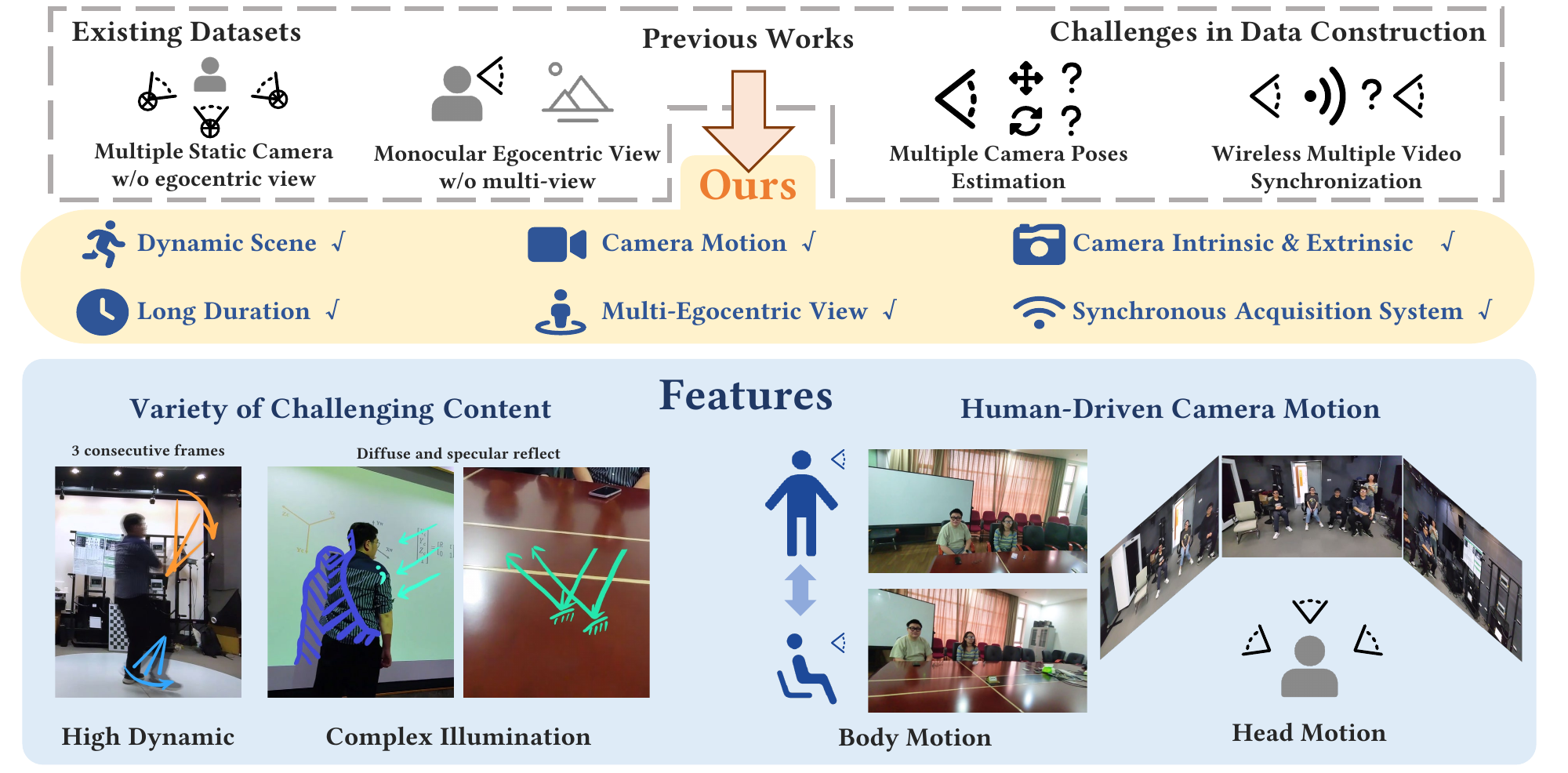}
  \caption{
  MultiEgo is the first multi-egocentric dynamic scene reconstruction dataset. The dataset provides essential data for reconstruction tasks, including synchronized egocentric videos and accurate camera pose annotations. It also features various challenges for reconstruction, such as human-driven camera motion, high-dynamic objects, and complex illumination.}
  \Description{ MultiEgo}
  \label{fig:teaser}
\end{teaserfigure}


\maketitle

\section{Introduction}

Free-viewpoint video (FVV), powered by dynamic scene modeling~\cite{n3dv,hypernerf,fast-nerf,hexplane,3d-4d}, represents a transformative leap in next-generation visual representation through its capacity for immersive real-time interaction. This paradigm unlocks unprecedented opportunities in entertainment, virtual reality, and human-computer interaction. Recent advances in novel view synthesis, such as 3D Gaussian Splatting~\cite{3DGS} (3DGS) have propelled dynamic scene reconstruction to new heights of efficiency and fidelity~\cite{3dgstream,4dgs,4dgc,hpc,varfvv,vrvvc,jointrf,spacetime,deformable-3dgs,gs4d,luiten2023dynamic}. Meanwhile, the rise of smart wearable devices (e.g. AR glasses) is redefining acquisition paradigms through lightweight, egocentric capture~\cite{egoexo4d,egocom,thu,hoi4d,kitchen,kitchen2,disney}. Unlike conventional fixed multi-camera systems~\cite{n3dv,Technicolor,Immersive,meetingroom}, multi-user wearable frameworks offer dual advantages: 1) eliminating interference of acquisition equipment while preserving natural participant behaviors, 2) enabling 4D social scene reconstruction via multi-view fusion~\cite{cmu1,cmu2,cmu3} in a more convenient way. These innovations pave the way for practical applications such as FVV meeting summaries and holographic concert recordings, heralding a new era of deployable dynamic scene reconstruction.

\begin{table*}[ht]
\centering
\caption{Comparison of MultiEgo and existing egocentric and 4D reconstruction Datasets.}
\label{tab:comparison}
    \begin{center}
    \begin{tabularx}{\textwidth}{X|ccc|cccc}
    \toprule[1.5pt]
     & \multicolumn{3}{c|}{Egocentric Datasets} & \multicolumn{3}{c}{4D reconstruction Datasets} & \cellcolor{gray!20}  \\ 
    \multirow{-2}{*}{\textbf{Character}}& 
    Ego4D~\cite{ego4d} & 
    Ego-Exo4D~\cite{egoexo4d} &
    Epic-Kitchens~\cite{kitchen,kitchen2,kitchen3} & 
    N3DV~\cite{n3dv}  & 
    D-NeRF~\cite{d-nerf} & 
    HyperNeRF~\cite{hypernerf} & 
    \cellcolor{gray!20}\multirow{-2}{*}{\textbf{Ours}} \\
    \hline 
    \rowcolor{gray!9}
    Dynamic scene      & \cmark & \cmark & \cmark & \cmark & \cmark    & \cmark  & \cellcolor{gray!20}\cmark  \\
    Egocentric view  & \cmark & \cmark & \cmark & \xmark & \cmark    & \cmark  & \cellcolor{gray!20}\cmark  \\
    \rowcolor{gray!9}
    Multi-perspective   & \cmark & \cmark & \xmark & \cmark & \xmark    & \xmark  & \cellcolor{gray!20}\cmark  \\
    Multi-egocentric view & \cmark & \xmark & \xmark & \xmark & \xmark    & \xmark  & \cellcolor{gray!20}\cmark  \\
    \rowcolor{gray!9}
    Camera poses provided & \xmark & \cmark & \xmark & \cmark & \cmark    & \cmark  & \cellcolor{gray!20} \cmark  \\
    \bottomrule[1.5pt]
    \end{tabularx}
    \end{center}
\end{table*}

However, existing datasets often exhibit critical limitations in multi-egocentric view reconstruction research: (1) Most dynamic scene datasets such as N3DV~\cite{n3dv}, employ fixed multi-camera setups with static viewpoints, whereas multi-egocentric capture is driven by natural human-driven motions; (2) While monocular moving viewpoint datasets like HyperNeRF~\cite{hypernerf} and D-NeRF~\cite{d-nerf}, lack multiple egocentric views, resulting in less information in reconstructing; (3) Egocentric-centric datasets such as HOI4D~\cite{hoi4d}, EPIC-Kitchens~\cite{kitchen,kitchen2,kitchen3}, EgoDex~\cite{egodex} primarily focus on human-object interactions or activity recognition tasks, including comprehensive benchmarks like Ego4D~\cite{ego4d}  and Ego-Exo4D~\cite{egoexo4d} which emphasize video understanding tasks rather than scene reconstruction. In a nut shell, no existing dataset simultaneously provides multi-person egocentric perspectives with synchronized pose estimations, which is the critical capability our dataset explicitly addresses.

In this paper, we present MultiEgo, the \textbf{first} multi-egocentric dynamic scene reconstruction dataset, addressing the limitations of existing datasets that primarily focus on fixed-camera and monocular egocentric settings. The dataset contains five dynamic scenes, including meetings, performances, and a presentation. Each was captured through five 1080p 30 FPS egocentric videos with accurate pose annotations. By making full use of the device, we addressed the synchronization challenges during recording. We conducted baseline evaluations and detailed analysis that validated the effectiveness of the dataset and provide insights for future task development.
Our dataset characteristics are compared with existing datasets, including egocentric datasets: Ego4D~\cite{ego4d}, Ego-Exo4D~\cite{egoexo4d}, EPIC-Kitchens~\cite{kitchen,kitchen2,kitchen3}, and 4D reconstruction datasets: N3DV~\cite{n3dv}, D-NeRF~\cite{d-nerf}, HyperNeRF~\cite{hypernerf}, as summarized in Table \ref{tab:comparison}.

Our contribution could be summarized as follows:

\begin{itemize}
\item We present the MultiEgo dataset, the first dataset for multi-egocentric dynamic scene reconstruction. The dataset contains five challenging daily social scenes, and each scene is composed of five strictly synchronized egocentric videos with accurate pose annotations. 
\item We design a customized multi-egocentric data acquisition system and a data process pipeline, enabling hardware-level synchronization across viewpoints and accurate pose estimation.
\item We conducted baseline evaluations on our dataset. The experimental results demonstrate its strong practical utility and effectiveness for dynamic scene reconstruction tasks.
\end{itemize}


\section{Related Works}

\noindent \textbf{Dynamic Scene Reconstruction Dataset.} With the advancement of computer vision technologies and data acquisition devices, numerous dynamic scene datasets have emerged~\cite{n3dv,d-nerf,Immersive,Technicolor,meetingroom,yoon2020dynamic}.. Representative dynamic datasets such as the N3DV~\cite{n3dv} dataset provides multi-view video recordings from fixed perspectives to capture dynamic scenes, where one or more performers perform various subtle interactions, such as cooking and talking. However, perspective coverage in N3DV is limited, and fixed viewpoints require specialized equipment for similar data collection. The HyperNeRF~\cite{hypernerf} dataset features scenes captured from a single moving viewpoint, recording brief actions such as breaking cookies or pouring liquids. Although data in HyperNeRF can be considered egocentric, it focuses on small-scale object-centric scenarios with only monocular observations.

\noindent \textbf{Egocentric Dataset.} In recent years, egocentric vision datasets have experienced rapid development~\cite{ego4d,egoexo4d,egocom,thu,hoi4d,kitchen,kitchen2,kitchen3,egodex,tu1,tu2,disney,2018charades,de2009guide,you2me,ineye,egohand}. Prominent examples include Ego4D ~\cite{ego4d} and EgoExo4D ~\cite{egoexo4d}, which feature large-scale data volume and diverse content modalities. However, these datasets typically provide only single-view egocentric perspectives and are primarily designed for video content understanding~\cite{delving,egoonly,DetectingADL} rather than scene reconstruction. With the emergence of multi-modal large models, several human-object interaction (HOI) datasets have been proposed, including HOI4D ~\cite{hoi4d}, Epic-Kitchens ~\cite{kitchen,kitchen2,kitchen3}, and EgoDex ~\cite{egodex}. While these datasets advance tasks like human behavior understanding, their design objectives and data characteristics make them unsuitable for direct application in dynamic scene reconstruction tasks. For example , EgoGaussian~\cite{egogaussian} attempts to reconstruct egocentric data on the HOI4D and Epic-Kitchens datasets but still requires excessively complex pre-processing to achieve viable results. This further demonstrates the critical value of purpose-designed multi-egocentric datasets for dynamic scene reconstruction.

\begin{figure*}[ht]
  \centering
  \includegraphics[width=\linewidth]{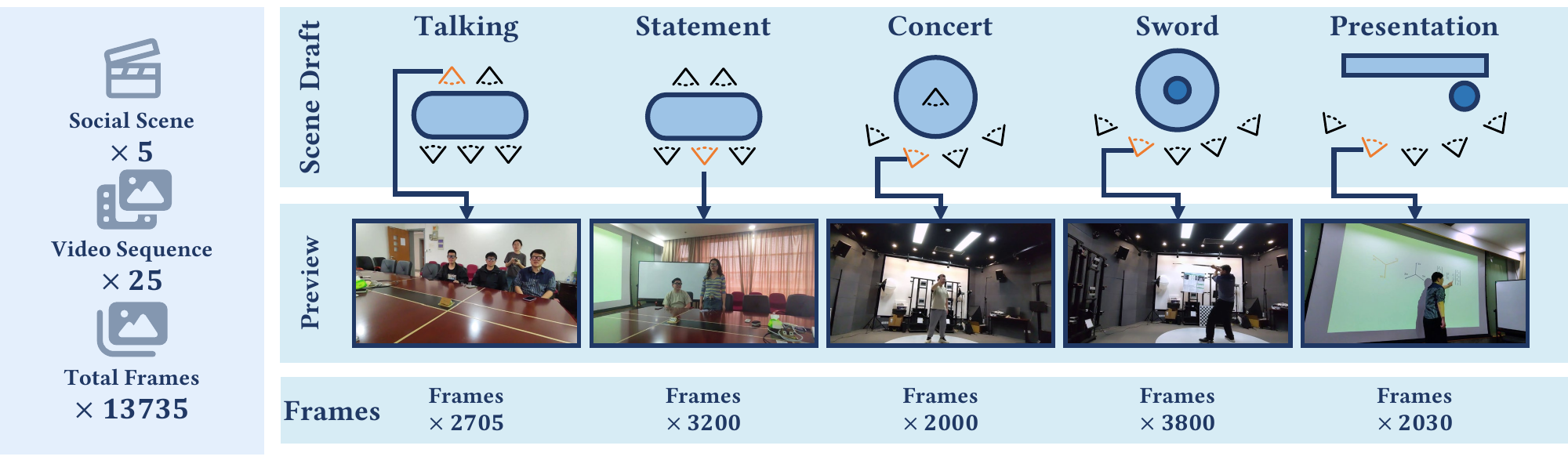}
  \caption{An overview of our MultiEgo dataset. There has 5 high-quality dynamic scenes in the dataset, each scene contains 5 egocentric views with accurate pose annotation. The total number of frames comes to 13,735. }
  \Description{the pipeline of data processing}
  \label{fig-content}
\end{figure*}

\section{MultiEgo Dataset}


\begin{figure*}[h]
  \centering
  \includegraphics[width=\linewidth]{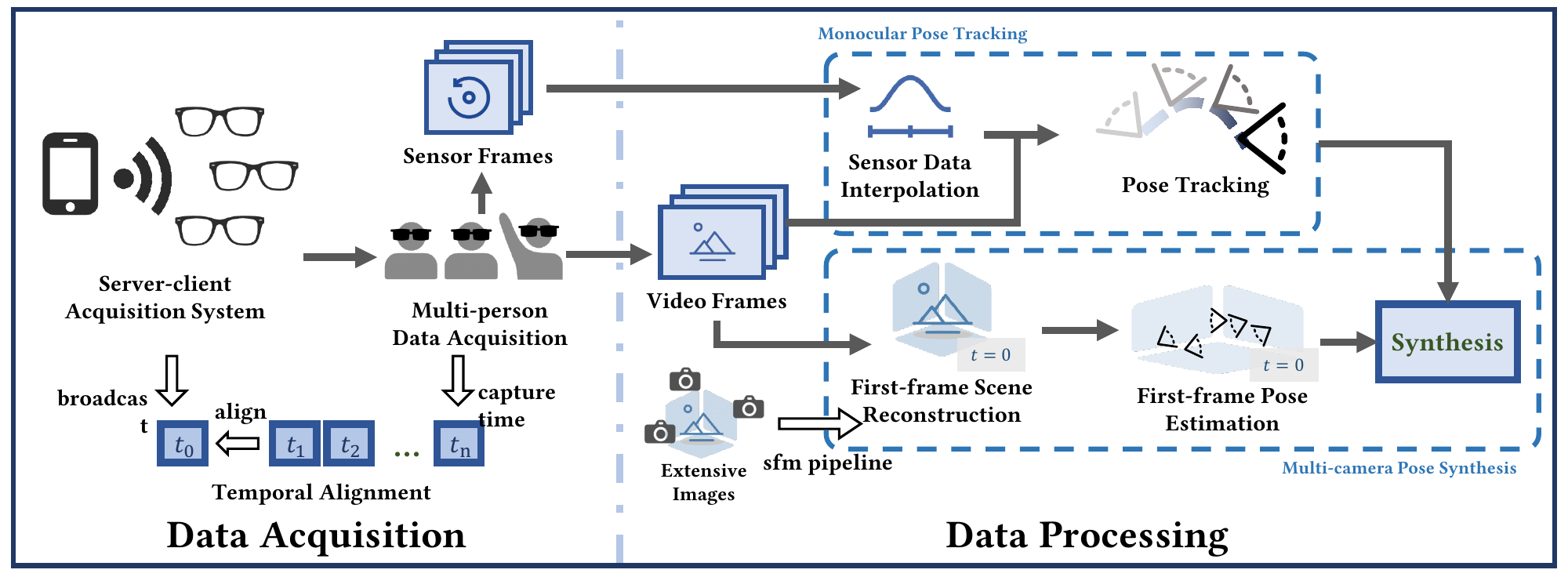}
  \caption{Pipeline of data acquisition and processing. We use a server-client system to handle synchronization, and respectively process time, monocular pose, poses of multiple views in the first frame. Finally we obtain the images and pose tracking in 4D domain.}
  \Description{the pipeline of data processing}
  \label{fig-process}
\end{figure*}

\subsection{Scene Overview}

To address the limitation that existing datasets are inapplicable to multi-view egocentric dynamic scene reconstruction, we present the MultiEgo dataset, the \textbf{first} dataset for multi-egocentric dynamic scene reconstruction. The dataset contains five multi-person social scenes, including meetings, performances, and a presentation. Each scene has five performers wearing AR glasses to provide authentic and reliable egocentric views. 
All participants signed informed consent forms authorizing the use of facial and other biometric data for academic research purposes.

The first scene called \textbf{\textit{talking}} involves a discussion meeting, in which performers take turns speaking. 
During this structured interaction, the performers tend to focus on the active speaker, resulting in systematic rotational camera movements through natural head rotations. This patterned motion enabled the system to capture comprehensive scene coverage despite the limited field-of-view of the individual cameras. 
Notably, although only five cameras were deployed, human-driven camera motions introduced diverse motion patterns that effectively enriched the observed information. 
In addition, the smooth wall and table surfaces of the meeting room induce extensive specular reflections from artificial lighting sources, accompanied by enhanced diffuse environmental reflections. This illumination phenomenon imposes significant challenges on reconstruction methods in determining the spectral properties of the surface reflectance while maintaining photometric consistency under complex lighting conditions.

The second scene called \textbf{\textit{statement}} involves a statement meeting, in which the performers take turns giving speeches while being required to stand up during their turns. During the transition from sitting to standing, we observed rapid translational movements caused by vertical body movement. 
In addition, after assuming the standing posture, certain performers exhibited natural exploratory behaviors characterized by horizontal head rotations, thereby enriching multi-perspective scene representations. 
Moreover, due to occlusion constraints in seated postures where performers cannot perceive others' lower body regions, the standing-up action effectively introduces new content in the scene, analogous to liquid pouring. This phenomenon imposes challenges on reconstruction algorithms.

The third scene called \textbf{\textit{concert}} features a standing performance paradigm with an actor and four audience members, all equipped with AR glasses. This configuration simulates real-world standing events such as concerts or public speeches, enabling dual-perspective analysis: (1) reconstructing actor's embodied performance from audience-centric viewpoints, and (2) capturing audience reactions through the actor's egocentric viewpoint. Although the actor constitutes the sole active visual source directed towards the audience, the frequent head/upper-body movements during performance generate dense spatial-temporal sampling of the scene, while the audience's motion magnitude remains small due to their stationary postures. This introduces valuable priors for novel view synthesis.

The fourth scene called \textbf{\textit{sword}} focuses on the dynamic action performance, where five audience members equipped with AR glasses observe an actor performing sword-wielding demonstrations using a long sword prop. This setup features sustained high-dynamic content characterized by complex 3D motion patterns: the limb movements of the actor exhibit rapid translational velocities, while the prop generates high angular velocities during slashing motions. These extreme motion characteristics pose significant challenges for high-speed motion reconstruction capabilities.

The fifth scene called \textbf{\textit{presentation}} simulates educational settings, for example lecture recordings, through a slide presentation task, in which five audience members equipped with the glasses observe an actor executing a presentation and delivering exaggerated gestures. This setup mimics real-world projection-based presentations by capturing two critical visual phenomena: (1) dynamic occlusion shadows formed when the performer's body blocks the projected light path and (2) illumination-induced chromatic variations that emerge on the performer's body as they move near the projected screen. The latter effect arises from the complex interplay between ambient lighting and high-luminance projector beams.

Each of the five scenes presents distinct technical challenges, including rapid object motions, high-speed camera movements, and specialized illumination conditions, such as specular surface reflections. These challenges represent critical prerequisites that must be addressed before advancing multi-egocentric reconstruction toward broad applications. In order for our dataset to enable rigorous evaluation of algorithm robustness under these challenging conditions, it explicitly incorporates scenarios specifically designed to assess resilience against complex visual phenomena. 


\subsection{Data Acquisition}\label{hardware}

\noindent \textbf{Hardware.}  In this paper, we select RayNeo X2 AR glasses for data collection, which is a consumer-grade AR device. The RayNeo X2 is equipped with a camera capable of recording 1080P resolution video at 30 frames per second. The device runs an Android operating system and supports WiFi communication. The official SDK provides real-time 3-degree-of-freedom rotational pose estimation from the built-in gyroscope sensor, which serves as an important foundation for camera pose estimation.

\noindent \textbf{Acquisition System.} To fully exploit the capabilities of the device, we developed a dedicated application system for data acquisition. The system integrates the functions of video capture, synchronized control, and sensor data collection. The system architecture adopts a client-server model: the client program runs on the AR glasses, continuously monitoring the WiFi channel to detect record/stop commands from the server. The server program runs on an external smartphone, establishing connections with the clients through a WiFi hotspot, and employs broadcast signals to synchronize recording start/end operations across multiple devices. 

Once the server program broadcasts the start signal, all clients initiate data acquisition within several microseconds, which can be regarded as practically simultaneous activation. This level of synchronization aligns with the requirements for multi-perspective scene reconstruction. During data acquisition, the client simultaneously captures visual and sensor data streams. The camera records 1080P video at 30 frames per second while the gyroscope provides rotation outputs at 50 Hz sampling rate. Notably, the client program records timestamps for each data frame (both video and sensor) in UTC with 100-nanosecond precision during collection, ensuring temporal synchronization across modalities and devices.

\subsection{Data Processing} \label{process}

\noindent \textbf{Video Post-processing.} To ensure visual consistency across multi-view egocentric videos, we applied Adobe Premiere Pro~\cite{pr} post-processing to standardize visual characteristics across all scenes. This included white balance calibration, exposure adjustment, and flicker removal from artificial lighting to eliminate sensor-specific color biases. Global exposure compensation and contrast adjustments further improved brightness uniformity while retaining shadow details.

 The pose estimation process consists of two components: monocular pose tracking and multi-camera pose synthesis. 
 
\noindent \textbf{Monocular Pose Tracking.} We applied a comprehensive set of state-of-the-art (SOTA) algorithms and engineering-validated classical methods to each camera's footage within every scene. Following the experimental benchmarks from Camerabench~\cite{camerabench}, we selected representative approaches including Anycam~\cite{anycam}, Mega-SAM~\cite{megasam}, CUT3R~\cite{cut3r}, MonST3R~\cite{monst3r}, PySLAM~\cite{pyslam}, covering both dynamic scene reconstruction (SfM) and simultaneous localization and mapping (SLAM) frameworks. 
We performed spherical linear interpolation~\cite{spherical} on the sensor's rotational quaternion data to estimate the rotational pose at each video frame capture moment. We perform data fusion~\cite{kalman} aligning all estimated trajectories with the interpolated sensor data. This alignment process utilized the 3-DoF rotational pose data from the official SDK to compute the relative 6-DoF camera poses for each view.

\begin{table*}[ht]
    \centering
    \caption{The quantitative results of our validation experiments. \bfseries{Bold}: Best result. \uline{Underline} : Second-best result.}
    \begin{tabularx}{\textwidth}{X|ccc|ccc|ccc}
    \toprule[1.5pt]
        \multirow{2}*{\textbf{Method}}& \multicolumn{3}{c}{\textbf{Talking}}& \multicolumn{3}{|c|}{\textbf{Statement}}& \multicolumn{3}{c}{\textbf{Concert}}\\ 
         & PSNR↑ & SSIM↑ & LPIPS↓ & PSNR↑ & SSIM↑ & LPIPS↓ & PSNR↑ & SSIM↑ & LPIPS↓ \\ \hline
        3DGStream~\cite{3dgstream} & 22.1335 & 0.7262 & 0.3382 & 20.4199 & 0.6711 & 0.3974 & 21.8715 & 0.7551 & 0.3074 \\ 
        Deformable-3DGS~\cite{deformable-3dgs} & 23.2186 & 0.8023 & 0.3358 & 21.3670 & 0.7731 & 0.3819 & 24.1020 & 0.8418 & \uline{0.2886} \\ 
        4DGaussian~\cite{4dgs} & \textbf{24.9863} & 0.8094 & 0.3353 & \textbf{24.0491} & \textbf{0.7894} & \textbf{0.3672} & \textbf{25.9235} & \textbf{0.8512} & 0.2953 \\ \hline
        4DGaussian~\cite{4dgs} w/ timestamp & \uline{24.9286} & \textbf{0.8102} & \uline{0.3319} & \uline{24.0174} & \uline{0.7875} & \uline{0.3701} & \uline{25.7934} & \uline{0.8490} & 0.2998 \\ 
        Deformable-3DGS~\cite{deformable-3dgs} w/ timestamp & 23.4073 & \uline{0.8058} & \textbf{0.3305} & 21.3836 & 0.7738 & 0.3801 & 24.1863 & 0.8427 & \textbf{0.2865} \\ \midrule[1.5pt]
        \multirow{2}*{\textbf{Method}} & \multicolumn{3}{c}{\textbf{Sword}}& \multicolumn{3}{|c|}{\textbf{Presentation}}& \multicolumn{3}{c}{\textbf{Average}} \\ 
         & PSNR↑ & SSIM↑ & LPIPS↓ & PSNR↑ & SSIM↑ & LPIPS↓ & PSNR↑ & SSIM↑ & LPIPS↓ \\ \hline
        3DGStream~\cite{3dgstream} & 24.0154 & 0.8284 & 0.2515 & 25.2535 & 0.8344 & 0.2866 & 22.7388 & 0.7630 & 0.3162 \\ 
        Deformable-3DGS~\cite{deformable-3dgs} & 21.2199 & 0.8603 & \uline{0.2358} & 25.2457 & 0.8872 & 0.2372 & 23.0306 & 0.8329 & \uline{0.2959} \\ 
        4DGaussian~\cite{4dgs} & \textbf{25.4828} & \textbf{0.8668} & 0.2671 & \uline{28.2414} & \uline{0.8994} & \uline{0.2265} & \textbf{25.7366} & \textbf{0.8432} & 0.2983 \\ \hline
        4DGaussian~\cite{4dgs} w/ timestamp & \uline{25.0993} & 0.8598 & 0.2778 & \textbf{28.2761} & \textbf{0.8995} & \textbf{0.2262} & \uline{25.6230} & \uline{0.8412} & 0.3012 \\ 
        Deformable-3DGS~\cite{deformable-3dgs} w/ timestamp & 20.9872 & \uline{0.8612} & \textbf{0.2345} & 24.5876 & 0.8857 & 0.2370 & 22.9104 & 0.8338 & \textbf{0.2937} \\ 
    \bottomrule[1.5pt]
    \end{tabularx}
    \label{exp table}
\end{table*}

\noindent \textbf{Multi-camera Pose Synthesis.} We instructed all performers to fixate on a common object in the first frame, establishing a static reference scene. For this initial static scene in the first frame, we captured extensive supplementary images from various angles to enrich the scene details and ensure robust reconstruction. All images were processed through the structure-from-motion pipeline in COLMAP~\cite{colmap} to obtain initial camera poses for all views. 
To maintain scale consistency across traces in monocular tracking, we selected an additional keyframe per view that contained content similar to the first frame but with noticeable translation. The displacement between these paired frames provided scale constraints to normalize translation components across all camera views.
After completing all the preparation steps, we integrated the poses of all frames within the same scene into absolute poses for scene reconstruction through global motion-guided pose synthesis. Specifically, we applied the following procedure:
1) Accumulated relative poses were anchored to the initial poses in COLMAP-reconstructed of each view,
2) Translation components were scaled using the displacement ratios from paired keyframes.
Experimental validation demonstrated that the resulting camera poses achieved sufficient accuracy for scene reconstruction tasks. Our data acquisition and processing pipeline is shown in Figure \ref{fig-process}.



\section{Experiment}

\subsection{Baselines}

Given the absence of prior literature on multi-egocentric dynamic scene reconstruction at the time of our dataset release, we therefore focus on adapting methods originally developed for general dynamic scene reconstruction, including deformation-filed-based methods 4DGaussian~\cite{4dgs} and Deformable-3DGS (D-3DGS)~\cite{deformable-3dgs}, and a streaming method 3DGSteam~\cite{3dgstream}.


Due to the lack of existing datasets, these baselines cannot be directly applied to our multi-egocentric dynamic scene reconstruction dataset. To evaluate our dataset's effectiveness with these baselines, we modified their data loading pipelines through a hard-coded approach that directly accesses per-frame images and camera poses. This modification preserves the core reconstruction mechanisms of the baselines while revealing their genuine performance in multi-egocentric dynamic scene reconstruction scenarios.

\subsection{Implementation Settings}

We adopt PSNR, SSIM, and LPIPS as quantitative evaluation metrics, which are widely used in scene reconstruction research. 
For all baselines, we adopt their officially recommended default settings. 
These methods generally initialize scene representations using observations from the first frame and predict motions of Gaussian primes, which works effectively for datasets with fixed or small-range camera movements. However, our dataset features large-range camera rotations that result in scene extents significantly exceeding the visible range captured in any single frame. When extensive new scene regions emerge across viewpoint changes, this can severely impact performance of the method with the initialization from a single frame. 
To mitigate this effect, we applied random point cloud initialization within the complete scene space for all baselines. In particular, we employ all available views to perform 3DGS~\cite{3DGS} pipeline on the randomly generated point cloud, thereby obtaining the full scene's Gaussian representation serving as the initialization for 3DGStream. Employing better initialization strategies tailored to our dataset may yield improved reconstruction outcomes.

\subsection{Experiment Results}

\begin{figure*}[t]
  \centering
  \includegraphics[width=\linewidth]{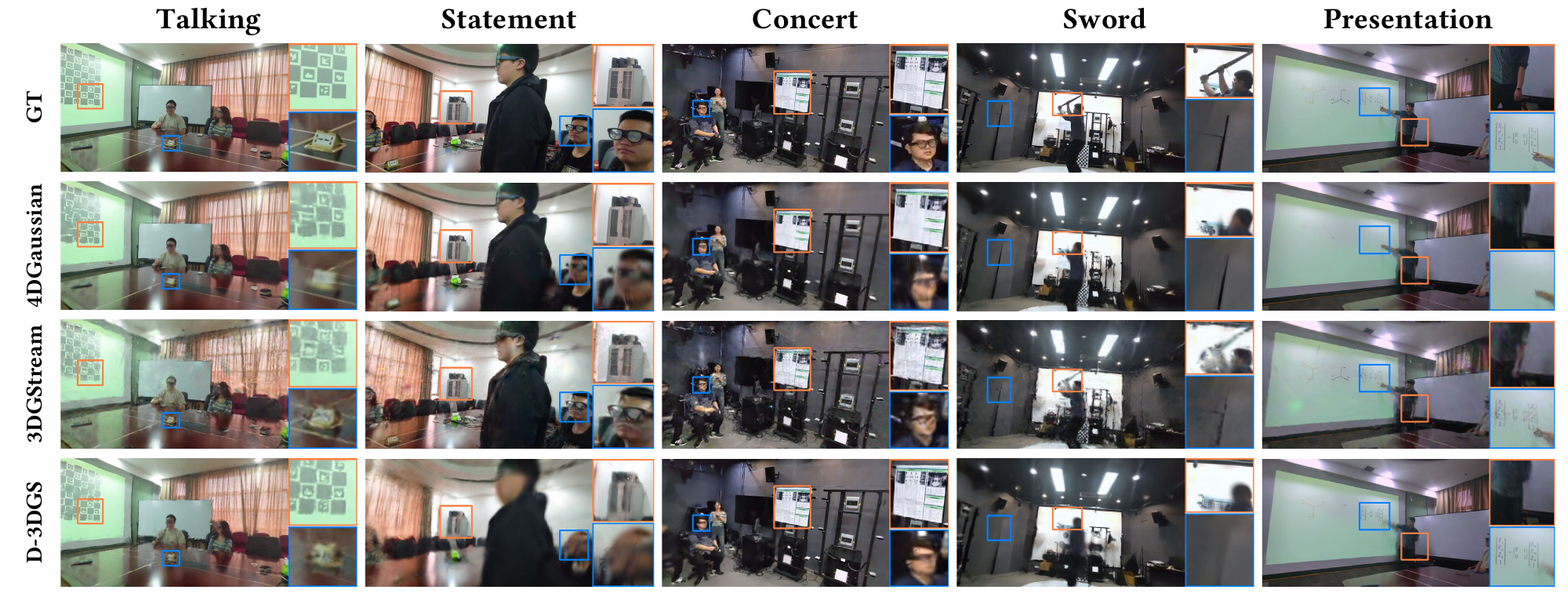}
  \caption{The visualization results from our validation experiments demonstrate that baselines employing different reconstruction strategies exhibit distinct characteristics.}
  \Description{The visualization results from our validation experiments demonstrate that baselines employing different reconstruction strategies exhibit distinct characteristics.}
  \label{fig-exp}
\end{figure*}

\noindent \textbf{Quantitative Results.} The quantitative experimental results of all selected baselines are summarized in Table \ref{exp table}. The result shows that the \textbf{\textit{sword}} and \textbf{\textit{presentation}} scene exhibit relatively smaller camera rotational motion magnitudes and more static background appearances, resulting in minimal novel background regions requiring reconstruction, which hence demonstrates relatively favorable quantitative results across methods. In contrast, the \textbf{\textit{statement}} scene presents large-scale camera rotations/translations combined with extensive specular reflections, making it the most technically challenging scenario among our dataset.

\noindent \textbf{Visualization Results.} The visualization results of our validation experiments are presented in Figure \ref{fig-exp}, where 4DGaussian demonstrates superior static scene reconstruction capabilities that align with its favorable quantitative metrics, but exhibits limitations in capturing high dynamic object. This performance discrepancy may stem from the MLP-based deformation field prediction mechanism in 4DGaussian, which prioritizes low-frequency information processing and consequently generates overly smoothed reconstructions. In contrast, 3DGStream produces reconstructions containing high-frequency noise but achieves the best preservation of high-dynamic details among all baselines, which is evident in sword prop reconstruction in the \textbf{\textit{sword}} scene. For multi-egocentric dynamic scene reconstruction, both high-quality background and accurate dynamic object are critical requirements. The trade-off between these two aspects constitutes a critical research challenge that must be addressed in future studies aiming at this task.


\noindent \textbf{Experiment About Timestamp.} As described in Section \ref{process}, we captured timestamps per frame. The collected timestamps closely align with the theoretical 30 FPS video capture intervals. Nevertheless, we conducted experiments incorporating actual timestamps as view-specific input in 4DGaussian and Deformable-3DGS, with results shown in "4DGaussian w/ timestamp" and "Deformable-3DGS w/ timestamp" of Table \ref{exp table}. Experimental results demonstrate that timestamps exert varying impacts across different scenes and methods, with performance improvements observed in some cases and degradations in others. Incorporating timestamps transforms the optimization from a single-scene estimation across five views to individual scene estimations per view with more dense time sampling. Although such distinctions exist, quantitative experiments demonstrate that this discrepancy is statistically insignificant.

\section{Conclusion}

We present MultiEgo, the first dynamic scene reconstruction dataset composed of multi-egocentric perspectives. Compared with previous dynamic scene reconstruction datasets and egocentric video collections, MultiEgo provides essential data elements for dynamic scene reconstruction tasks, including accurate camera pose annotations and synchronized temporal alignment. The dataset incorporates various challenges commonly encountered in multi-egocentric dynamic reconstruction scenarios, such as high dynamic objects and complex lighting conditions. We provide comprehensive descriptions of the dataset's contents. In validation experiments, we demonstrate the effectiveness of the data set through several baselines. In addition, we conducted additional studies to evaluate the impact of directly captured frame timestamps on reconstruction performance.


\begin{acks}
This work was partly supported by the NSFC62431015, Science and Technology Commission of Shanghai Municipality No.24511106200, the Fundamental Research Funds for the Central Universities, Shanghai Key Laboratory of Digital Media Processing and Transmission under Grant 22DZ2229005, 111 project BP0719010.
\end{acks}

\bibliographystyle{ACM-Reference-Format}
\bibliography{cite}










\end{document}